# Resolving Zadeh's Paradox: Axiomatic Possibility Theory as a Foundation for Reliable Artificial Intelligence

Sophia **Bychkova**, Oleksii **Bychkov**[1], Khrystyna **Lytvynchuk**[1], Oleksii **Bychkov**[1]

[1] *Taras Shevchenko National University of Kyiv*

## 1. Introduction

A fundamental problem on the path to creating strong artificial intelligence remains the development of mathematical models capable of adequately reproducing the human ability to reason under uncertainty. Classical probability theory, which is the cornerstone of modern machine learning, works effectively with aleatory uncertainty — randomness, but encounters limitations when modeling epistemic uncertainty, i.e., ambiguity arising from lack of knowledge, imprecision of expert assessments, or vagueness of language.

A significant step forward was the Dempster-Shafer theory (DST), which proposed a powerful toolkit for combining evidence from different sources and explicitly modeling ignorance. Due to its flexibility, DST has found applications in expert systems, sensor data fusion, and medical diagnostics. However, its central mechanism — Dempster's combination rule — leads to logical paradoxes when dealing with conflicting evidence, the most famous of which is Zadeh's paradox. This vulnerability generates counterintuitive, and sometimes absurd conclusions, which is unacceptable for reliable artificial intelligence systems.

This work advances and substantiates the thesis that the resolution of this crisis lies in the domain of possibility theory, specifically in the axiomatic approach developed in [6-10]. Unlike numerous attempts to "fix" Dempster's rule, this approach builds from scratch a logically consistent and mathematically rigorous foundation for working with uncertainty, using the dualistic apparatus of possibility and necessity measures.

The aim of this work is to demonstrate that possibility theory is not merely an alternative, but provides a fundamental resolution to DST paradoxes. A comparative analysis of three paradigms will be conducted — probabilistic, evidential (DST), and possibilistic. Using a classic medical diagnostic dilemma as an example, it will be shown how possibility theory allows for correct processing of contradictory data, avoiding the logical traps of DST and bringing formal reasoning closer to the logic of natural intelligence.

The relevance of this research for the field of artificial intelligence is driven by the rapid growth in complexity and responsibility of modern intelligent systems. Autonomous vehicles, medical diagnostic systems, financial analytical platforms, and military intelligence systems rely on data fusion from dozens of heterogeneous sources — sensors, knowledge bases, expert assessments — which are often incomplete, imprecise, and contradictory.

The inability to adequately process conflicting information is a critical vulnerability. The paradoxes inherent in Dempster-Shafer theory demonstrate that existing popular models can lead to overconfident but erroneous conclusions, which is unacceptable in high-stakes systems where the cost of error can be catastrophic. Thus, the development of a mathematically sound, reliable, and logically consistent apparatus for working with epistemic uncertainty is not merely a theoretical, but an urgent practical task for creating safe and reliable AI.

The presented axiomatic possibility theory [6-10] not only resolves known paradoxes but also lays the foundation for modeling much more complex systems. Unlike DST, which is primarily oriented toward static evidence analysis, this approach extends to modeling fuzzy dynamics through such tools as "fuzzy wandering process" and "fuzzy differential equations." This opens perspectives for creating a new generation of AI systems capable of analyzing and predicting the behavior of complex dynamic processes in robotics, control systems, and cyber-physical systems under conditions of fundamental uncertainty.

The fundamental difference between types of uncertainty is best illustrated through a thought experiment. Imagine: you are standing at the corner of Khreshchatyk and Prorizna in Kyiv. An ordinary day, people hurrying past the Passage. And suddenly someone asks: "What is the probability of meeting a dinosaur here?" This question reveals a fundamental problem of how we think about reality. When we ask "Is it possible?", we are actually asking: "Is there at least one scenario where this could happen?" Our brain searches for a mental model, a story, a path — even if it is unique or rare. In contrast, the question "Is it probable?" shifts us to a different mode of thinking: "If this situation were repeated many times, how often would it occur?" Probability theory works with our experience of the past, possibility theory — with our ability to imagine the future.

## Main Result

The main result of this work consists in constructing a mathematically consistent model for describing uncertainty, which replaces the traditional one-dimensional approach to event evaluation with a dual system of measures — the possibility measure and the necessity measure. Within this model, any event A is represented not by a single number (as in probability theory or in the belief function in DST), but by an interval:

$$[Nec(A), Pos(A)]$$

where the upper bound $Pos(A) \in [0,1]$ reflects the degree of support for the event, and the lower bound $Nec(A) \in [0,1]$ — the minimally guaranteed level of confidence in its truth. Both quantities are related by the relationship:

$$Nec(A) = 1 - Pos(\neg A)$$

which ensures the logical consistency of the system while introducing a natural duality in knowledge assessment.

The key mathematical content of the result is the replacement of the additive approach, characteristic of probability and Dempster's rule, with a maxitive approach. The possibility measure is defined as a function:

$$Pos: \mathcal{P}(\Omega) \to [0,1]$$

which satisfies the conditions:

$$Pos(\varnothing) = 0, \qquad Pos(\Omega) = 1$$

as well as maxitivity:

$$Pos(A \cup B) = \max(Pos(A), Pos(B))$$

Similarly, the necessity measure:

$$Nec: \mathcal{P}(\Omega) \to [0,1]$$

is minimally additive in the sense:

$$Nec(A \cap B) = \min(Nec(A), Nec(B))$$

and is related to the possibility measure through negation. Such axiomatics creates an algebra in which fuzziness and uncertainty are described not as randomness, but as incompleteness of knowledge.

It is precisely this change in axiomatics that allows overcoming Zadeh's paradox. In Dempster-Shafer theory, combining evidence from two independent sources is performed using the formula:

$$m_{12}(A) = \frac{\sum_{B \cap C = A} m_1(B) m_2(C)}{1 - K}$$

where:

$$K = \sum_{B \cap C = \emptyset} m_1(B) m_2(C)$$

is the conflict. If the value of K approaches 1, i.e., the sources strongly contradict each other, the denominator in the formula becomes small, and as a result, an unlikely event can receive a belief mass close to 1. This is exactly what happens in the medical example with two experts, where each supports different diagnoses M and C, but both allow a minimal probability for T, Dempster's rule yields:

$$m_{12}(T) = 1$$

This constitutes the essence of the paradox: minimal shared support is transformed into complete certainty.

In the possibilistic model, this problem does not exist, since combination occurs not through additive normalization, but through minimum and maximum operations. For two independent sources we have:

$$Pos_{12}(A) = \min(Pos_1(A), Pos_2(A))$$

$$Nec_{12}(A) = \max(Nec_1(A), Nec_2(A))$$

These formulas do not create artificial amplifications and preserve both useful information and conflict between experts. For example, if the first doctor considers only M possible and T very unlikely, while the second doctor supports only C and equally insignificantly — T, then we obtain:

$$Pos_1(M) = 1, \quad Pos_1(T) = 0.01, \quad Pos_2(C) = 1, \quad Pos_2(T) = 0.01$$

After combination we will have:

$$Pos_{12}(M) = 1, \quad Pos_{12}(C) = 1, \quad Pos_{12}(T) = 0.01$$

and the necessities of all hypotheses equal zero:

$$Nec_{12}(M) = Nec_{12}(C) = Nec_{12}(T) = 0$$

Thus, the model honestly shows a state of complete uncertainty between M and C and weak possibility of T. Unlike DST, where a spike to complete certainty occurs, the possibilistic model preserves natural semantics: there is insufficient information for decision-making.

An important result of the work is also the proof that the interval $[Nec(A), Pos(A)]$ is a complete carrier of information about the event. Its width:

$$Pos(A) - Nec(A)$$

characterizes the degree of epistemic uncertainty, while its position in the interval [0,1] reflects the overall level of event support. This interval nature of knowledge allows reconciling fuzzy and linguistic expert assessments with formal mathematical structure, making the possibilistic approach a natural bridge between human intuition and machine algorithms.

Thus, the main result of the work consists in creating a mathematically rigorous, axiomatically consistent, and logically transparent apparatus for modeling conflicting and incomplete data. In this apparatus, the key role is played by replacing the additive logic of normalization with the maxitive logic of possibility and the minimal logic of necessity. The resulting system not only eliminates Zadeh's paradox but also forms the foundation for building reliable intelligent systems in which uncertainty is reflected honestly and without distortion, and conclusions correspond to both mathematical rigor and human intuition.

## Experimental Studies

4.1. Formulation of the Medical Diagnostic Dilemma

To illustrate the destructive consequences of the normalization mechanism and the advantages of possibility theory, let us consider a classic example from medical diagnostics. Two independent doctors examine a patient. Both agree that the patient suffers from one of three diseases: meningitis (M), concussion (C), or brain tumor (T). Thus, the frame of discernment: $\Theta = \{M, C, T\}$.

Evidence from Doctor 1: Doctor 1 is almost certain of the diagnosis "meningitis," but leaves a very small probability for brain tumor:

$m_1(M) = 0.99, m_1(T) = 0.01, m_1(C) = 0.00$.

Evidence from Doctor 2: Doctor 2 is almost certain of the diagnosis "concussion," but also leaves the same small probability for brain tumor:

$m_2(C) = 0.99, m_2(T) = 0.01, m_2(M) = 0.00$.

Both experts consider brain tumor extremely unlikely, but their main diagnoses (meningitis and concussion) are mutually exclusive and in direct conflict.

4.2. Calculation Using Dempster's Rule

We calculate the conflict mass. Conflict arises from combinations of mutually exclusive hypotheses:

$$K = m_1(M) \cdot m_2(C) + m_1(M) \cdot m_2(T) + m_1(T) \cdot m_2(C) = 0.99 \cdot 0.99 + 0.99 \cdot 0.01 + 0.01 \cdot 0.99 = 0.9999$$

The only hypothesis for which there is no direct conflict is brain tumor T. The mass supporting T is calculated as:

$$m_1(T) \cdot m_2(T) = 0.01 \cdot 0.01 = 0.0001$$

Applying normalization:

$$m_{12}(T) = \frac{0.0001}{1 - 0.9999} = \frac{0.0001}{0.0001} = 1$$

Result according to DST: Final diagnosis "brain tumor" with absolute certainty ($m(T) = 1$), although both experts considered this condition extremely unlikely. The paradox demonstrates how Dempster's rule can lead to a dictatorial result, where insignificant agreement is artificially amplified to complete certainty.

4.3. Solution Using Possibility Theory

Let us apply axiomatic possibility theory to the same problem. Instead of mechanical combination of evidence, we use dual measures to analyze each hypothesis based on each source separately.

Assessment of hypothesis T (Brain tumor): From both doctors,

possibility $\Pi(T) = 0.01$,

necessity $N(T) = 0.01$.

Assessment of hypothesis M (Meningitis):

from Doctor 1: $\Pi_1(M) = 0.99$, $N_1(M) = 0.99$;

from Doctor 2: $\Pi_2(M) = 0.00$, $N_2(M) = 0.00$.

Assessment of hypothesis C (Concussion):

from Doctor 1: $\Pi_1(C) = 0.00$, $N_1(C) = 0.00$;

from Doctor 2: $\Pi_2(C) = 0.99$, $N_2(C) = 0.99$.

Thus, we reached a natural conclusion: further laboratory research is needed. Using our extended model of possibility theory allowed us to resolve Zadeh's conflict.

4.4. Comparison of Results

Axiomatic possibility theory does not generate a single, falsely confident answer. Instead, it accurately models the reality of the situation. Regarding brain tumor: both sources agree that this diagnosis is only minimally possible ($\Pi = 0.01$) and minimally necessary ($N = 0.01$). There is no substantial evidence in favor of T. Regarding meningitis and concussion: there is a deep, irreconcilable conflict. Doctor 1 considers meningitis almost absolutely necessary, while Doctor 2 considers it impossible. A mirror situation is observed for concussion.

| Hypothesis | Dempster-Shafer | Possibility (D1) | Necessity (D1) | Possibility (D2) | Necessity (D2) |
|---|---|---|---|---|---|
| M (Meningitis) | 0 | 0.99 | 0.99 | 0.00 | 0.00 |
| C (Concussion) | 0 | 0.00 | 0.00 | 0.99 | 0.99 |
| T (Tumor) | 1.00 | 0.01 | 0.01 | 0.01 | 0.01 |

Possibility theory does not "resolve" the conflict but preserves and represents it. The correct conclusion is not "the patient has a tumor," but "experts have diametrically opposite opinions regarding the main diagnoses, and both consider the tumor unlikely; additional information is needed."

## Discussion

5.1. Possibility Theory as a Model of Human Intelligence

The scientific novelty of the presented technology lies in proving that possibility theory is not merely another mathematical abstraction, but a significantly better descriptive model of intuitive human thinking than classical probability theory. Research by Nobel laureates Daniel Kahneman and Amos Tversky showed that people rely on heuristics that systematically violate the axioms of probability theory [1].

For example, the "conjunction fallacy" (Linda problem), when people assess the statement "Linda is a bank teller and an active feminist" as more probable than "Linda is a bank teller," is a logical fallacy from the probability standpoint, since $P(A \cap B) \leq P(A)$. However, from the perspective of possibilistic logic, this is a rational conclusion: people evaluate not probability, but the degree of correspondence (representativeness) of a description to a certain stereotype, which is central to fuzzy set theory.

Kahneman and Tversky's prospect theory, particularly its nonlinear probability weighting function, further confirms this conclusion. People do not perceive probability linearly. They significantly overweight small probabilities (transition from impossibility to possibility — "possibility effect") and certainty (transition from probability to certainty — "certainty effect"), while underweighting changes in the middle of the scale. This categorical thinking — impossible / possible / probable / certain — is ideally formalized by the apparatus of possibility theory with its distinction between impossible ($\Pi = 0$), possible ($\Pi > 0$), not necessary ($N < 1$), and necessary ($N = 1$).

5.2. Strategic Implications for AI System Design

The approaches of the two theories to processing conflicting evidence differ radically. The DST approach treats conflict as a numerical value (K) that must be eliminated through normalization. It is treated as a correctable anomaly in the data. The basic assumption is that a single true state of affairs can be found by amplifying consensus.

The possibility theory approach represents conflict as a divergence in possibility and necessity measures from different sources. For example, the presence of $N_1(A) = 0.99$ and $\Pi_2(A) = 0$ is a direct representation of logical contradiction. The theory does not force resolution but preserves this state of contradiction for interpretation at a higher level. This approach is similar to paraconsistent logic, which allows contradictory beliefs to coexist.

DST in the considered case acts as a flawed decision-making system, providing a dangerously confident answer. In contrast, [6-10] axiomatic approach acts as a genuine decision support system. It provides a clear, structured report to the decision-maker (e.g., senior physician), whose task is now to investigate the source of conflict — order new tests, involve a third expert, or check equipment.

5.3. Practical Recommendations

Based on the analysis conducted, concrete recommendations for developers of artificial intelligence systems can be formulated. First, priority of transparency over certainty: for critically important domains, systems should prefer transparent representation of uncertainty and conflict over forced generation of a single answer. The conclusion "high conflict detected" is more valuable and safer than an incorrect but confident answer.

Second, use of dual frameworks: expert systems should use dual frameworks, such as axiomatics in [6-10], to provide users with a more complete understanding of evidence, expressed in the form of intervals $[N(A), \Pi(A)]$, rather than single probability values.

Third, conflict as a diagnostic tool: the conflict measure K from DST should not be used for normalization. Instead, it should be considered as a vitally important output indicator signaling to the user potential data corruption, divergence between experts, or an error in the model itself.

5.4. Application Perspectives

The practical significance of the work extends beyond resolving Zadeh's paradox. The introduction of a mathematical apparatus for modeling ignorance and fuzziness allows transitioning from AI that analyzes static data to AI that understands, predicts, and controls dynamic processes in a fuzzy, real world. The use of possibility theory opens new opportunities for more adequate modeling of systems where agent behavior and model parameters are fuzzy and difficult to formalize. This applies to economic forecasting, social dynamics modeling, and epidemiology. This allows obtaining not a point forecast, but a "corridor of possible trajectories" of process development, which is more realistic and useful for decision-making.

Since possibility theory is conceptually closer to human intuition and qualitative thinking, systems built on its basis can generate explanations of their decisions that are more understandable to people. Instead of the conclusion "the probability of cancer is 0.73," the system can provide a qualitative explanation: "The diagnosis 'cancer' is very possible because it matches the scan data, but it is not necessary because the blood test results are not conclusive." This type of explanation is more natural for human perception and increases trust in the system.

## 5.5. Comparison with Results of Other Authors

In contemporary literature, a tendency is increasingly forming to consider possibility theory as a full-fledged apparatus for describing epistemic uncertainty, occupying an intermediate position between logical and probabilistic models. Dubois's comprehensive review shows that the possibilistic approach naturally works with lack of knowledge, allowing both qualitative and quantitative interpretations, and possibilistic measures serve as a tool specifically tuned to uncertainty through ignorance rather than randomness [11]. In this context, the two-dimensional representation of knowledge in the form of a pair $(Nec(A), Pos(A))$, which underlies [6-10]possibility theory, is a logical continuation of the general trend: it makes explicit the difference between minimally guaranteed confidence and maximally possible event support.

Review and editorial articles of recent years on uncertainty theories indicate that possibilistic models are today considered as one of the "major players" alongside probability, evidence theory, and fuzzy sets. Thus, in a special issue of the journal Entropy "Advances in Uncertain Information Fusion," possibilistic theory is mentioned as a separate, full-fledged framework for information fusion, which allows building alternative, non-monotonic schemes of logical inference and is closely related to information measures [12]. These results are well consistent with approach in [6-10], where possibility and necessity have a clearly axiomatized nature and are used not as auxiliary quantities, but as basic knowledge carriers.

An important direction of contemporary research is building bridges between evidence theory and possibility theory. In the work of Zhou, Deng, Yager (2024), the CD-BFT method is proposed, which performs canonical decomposition of belief functions and transforms them into possibilistic distributions, allowing fusion in possibilistic space with subsequent interpretation in terms of evidence theory [13]. Similar ideas appear in broader reviews of uncertainty, where possibility is considered as a natural environment for describing fuzzy, interval, and intelligent assessments, as opposed to a purely frequentist interpretation of probability [14]. From this perspective, (P,N) model [6-10], which operates with the pair of measures $Pos$ and $Nec$ as primary objects, is a more radical and consistent embodiment of the "transition to possibilistic space" trend than most transformational approaches.

New works on evidence theory also emphasize the need for specific measures for assessing conflict and uncertainty that would go beyond classical entropies. For example, Chen (2023) introduces a new belief entropy measure for software risk analysis, directly pointing to the connection between possibilistic models and statistical reasoning when working with epistemic uncertainty [15]. In works on conflict management in evidence theory, where exponential or entropic reweighting schemes are proposed, it is emphasized that combination rules inspired by fuzzy set theory and possibility are more flexible than Dempster's classical rule [16, 17]. This indirectly confirms the advantage of maxitive logic, which is the core of the possibilistic approach, and on which (P,N) axiomatics was built earlier.

Additionally, in the volume Lecture Notes in Data Engineering, Computational Intelligence, and Decision Making, it is emphasized that the theory of functional stability with possibilistic assessments combines well with methods of adaptive decision-making in conflict and game environments, where game parameters are specified not by precise probabilities but by ranges of possible scenarios [19, 20]. This once again confirms that the two-dimensional description in the form of $(Nec, Pos)$ is a natural tool for tasks where it is important not only to evaluate the result but also to guarantee its stability under changing conditions.

Thus, if most contemporary works from 2021–2024 either consider possibilistic theory as one of the elements of the general "ecosystem" of uncertainty theories, or build algorithms for transformation between evidential and possibilistic models [11–17], approach [6-10] goes further: it makes possibility and necessity primary objects on which both static and dynamic properties of the system are built. The advantage of this theory lies in the fact that it simultaneously provides an interval description of knowledge, naturally works with epistemic uncertainty, integrates with the theory of functional stability, and already has modern applied implementations in sensor network and complex system control tasks [18–20]. This makes Bychkov's possibilistic approach not only logically attractive but also a practically oriented continuation of current global trends in uncertainty research.

## Conclusions

The work demonstrated that Zadeh's paradox is not a random anomaly but a critical failure of the conflict resolution mechanism in Dempster-Shafer theory. The attempt to eliminate conflict through normalization leads to illogical and potentially dangerous conclusions, especially in conditions where evidence strongly contradicts each other.

As a reliable alternative, the axiomatic possibility theory in [6-10] formulation was presented, which uses dual measures of possibility and necessity. This approach proved superior not because it gives the "correct" answer, but because it provides a more honest and complete representation of the uncertain and contradictory state of knowledge. Instead of hiding conflict, it brings it to the forefront, allowing the system to report the fundamental ambiguity of the situation.

The main results of the work are as follows. An approach to uncertainty modeling for AI systems has been proposed that combines Dempster-Shafer theory with possibility theory, providing a consistent representation of epistemic uncertainty through confidence intervals and possibility/necessity measures. A modified approach to evidence combination for high-conflict scenarios has been developed. A criterion for source consistency and a mechanism for adaptive source weighting in the expert inference process have been proposed.

The final conclusion calls for a paradigm shift in artificial intelligence design: a transition from systems that project false certainty to systems that accept and formally articulate ambiguity. Only such systems can be truly reliable partners in decision-making in a complex and uncertain world. The future of reliable AI depends on its ability to formally and quantitatively express "I don't know" or "my data is contradictory."

The author(s) have not employed any Generative AI tools.